\documentclass[twoside,11pt]{article}

\usepackage{jmlr2e}
\usepackage{enumitem}
\usepackage{amsmath}
\usepackage{amssymb}
\usepackage{hyperref}
\usepackage[htt]{hyphenat}
\usepackage{clipboard}

\ShortHeadings{\texttt{YAMLE}: Yet Another Machine Learning Environment}{Ferianc, Rodrigues}
\firstpageno{1}

\author{\name Martin Ferianc$^{1}$ \email martin.ferianc.19@ucl.ac.uk \\
\name Miguel Rodrigues$^{1}$ \email m.rodrigues@ucl.ac.uk \\
\addr $^{1}$Department of Electronic and Electrical Engineering,\\
University College London,
London, United Kingdom}

\begin{document}

\title{\texttt{YAMLE}: Yet Another Machine Learning Environment}


\maketitle

\begin{abstract}
\texttt{YAMLE}: Yet Another Machine Learning Environment is an open-source framework that facilitates rapid prototyping and experimentation with machine learning (ML) models and methods. 
The key motivation is to reduce repetitive work when implementing new approaches and improve reproducibility in ML research. 
\texttt{YAMLE} includes a command-line interface and integrations with popular and well-maintained PyTorch-based libraries to streamline training, hyperparameter optimisation, and logging. 
The ambition for \texttt{YAMLE} is to grow into a shared ecosystem where researchers and practitioners can quickly build on and compare existing implementations. 
Find it at: \url{https://github.com/martinferianc/yamle}.
\end{abstract}


\section{Introduction}\label{sec:introduction}

The success of machine learning (ML) in recent years has been driven by the availability of open-source software such as \texttt{Caffe}~\citep{jia2014caffe}, \texttt{Keras}~\citep{chollet2015keras}, \texttt{Theano}~\citep{al2016theano},  \texttt{TensorFlow}~\citep{45381}, \texttt{PyTorch}~\citep{paszke2017automatic}, \texttt{PyTorch Lightning}~\citep{falcon2019pytorch} or \texttt{Jax}~\citep{frostig2018compiling} among others. 
These libraries have enabled researchers and practitioners to build and experiment with ML quickly.
Imagine a researcher who wants to experiment with a new ML model or method.
Researchers need to compare their models or methods to state-of-the-art approaches to determine whether they are better than the existing ones.
They must implement the existing models or methods and compare them on the same datasets and tasks.
Therefore, they implement the existing models or methods and all the boilerplate code, such as data loading, preprocessing, logging, hyperparameter optimisation, evaluation, and the connections between all the pipeline's components.
By the nature of reimplementing complicated models or methods, this process often results in disparate implementations and a lack of standardisation, hindering reproducibility and comparison of results~\citep{semmelrock2023reproducibility}.
But is the reimplementation of all the boilerplate code, its connectivity, and the models or methods needed?

This paper introduces \texttt{YAMLE}: Yet Another Machine Learning Environment -- an open-source generalist customisable experiment environment with boilerplate code already implemented for rapid prototyping with ML models and methods. 
The main features of the environment are summarised as follows:

\begin{figure}[t]
    \centering
    \includegraphics[width=1\linewidth]{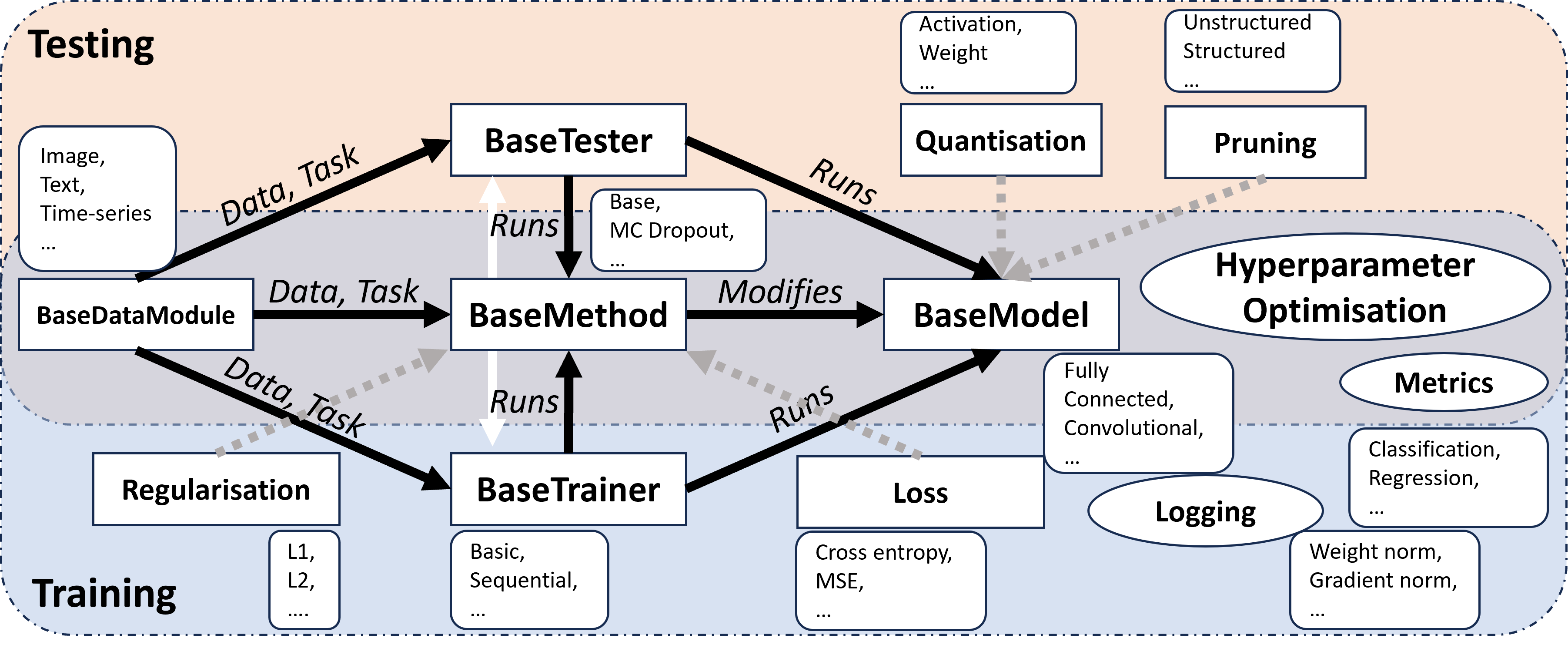}
    \vspace{-0.5cm}
    \caption{The overview of the environment's design. 
    It consists of three main components - \texttt{BaseDataModule}, \texttt{BaseModel} and \texttt{BaseMethod} managed by \texttt{BaseTrainer/BaseTester}. The \texttt{BaseDataModule} is responsible for downloading, loading and preprocessing data. 
    The \texttt{BaseModel} defines the model's architecture. 
    If necessary, the \texttt{BaseMethod} changes the model and defines the training, validation and test steps. 
    The \texttt{BaseTrainer/Tester} groups the \texttt{BaseDataModule}, \texttt{BaseModel} and \texttt{BaseMethod} together with \texttt{Logging}. 
    The whole training and testing can be overseen by \texttt{Hyperparameter Optimisation}. 
    Additional components which actively change the training and can be defined by the user are \texttt{Regularisation}, \texttt{Quantisation} and \texttt{Pruning}.}
    \vspace{-0.5cm}
    \label{fig:design}
\end{figure}

\begin{itemize}[leftmargin=*, topsep=-0.12cm, itemsep=-0.1cm]
    \item \textbf{Modular Design} - The environment is divided into three main components - data, models, and methods - which are infrastructurally connected but can be independently modified and extended.
    The goal is to write a method or a model and then seamlessly use it across different models or methods across different datasets and tasks.
    \item \textbf{Command-line Interface} - The environment includes a command-line interface for easy configuration of all hyperparameters and training of models.
    \item \textbf{Hyperparameter Optimisation} - The environment is integrated with \texttt{syne-tune}~\citep{salinas2022syne} for hyperparameter optimisation.
    \item \textbf{Logging} - The environment is integrated with TensorBoard~\citep{45381} for logging and visualisation of training, validation and test metrics, supported by \texttt{torchmetrics}~\citep{torchmetrics} and \texttt{PyTorch Lightning}~\citep{falcon2019pytorch}.
    \item \textbf{End-to-end Experiments} - \texttt{YAMLE} enables end-to-end experiments from data preprocessing to model training and evaluation.
    All settings are recorded for reproducibility.
\end{itemize}

\section{Core Components and Modules}\label{sec:core-components-modules}

\texttt{YAMLE} is built on PyTorch and \texttt{PyTorch Lightning}. 
In contrast to other environments, \texttt{YAMLE} opted for PyTorch given its popularity and ease of use~\citep{tidjon2022empirical}, making it a suitable choice for a general-purpose ML environment. 
The framework relies on \texttt{torchmetrics}~\citep{torchmetrics} for evaluation metrics and \texttt{syne-tune}~\citep{salinas2022syne} for hyperparameter optimisation.
Overall, the environment's purpose is to provide an ecosystem for rapid prototyping and experimentation by modifying the core components - \texttt{BaseDataModule}, \texttt{BaseModel}, and \texttt{BaseMethod} - as shown in Figure~\ref{fig:design}, and then used across different models, methods, datasets, and tasks.

\subsection{\texttt{BaseDataModule}}\label{sec:core-components-modules:datamodule}

The \texttt{BaseDataModule}, defined in \path{yamle/data/datamodule.py}, is responsible for downloading, loading, and preprocessing data. 
It defines the task, e.g., classification or regression, to be solved by the \texttt{BaseMethod} and \texttt{BaseModel} and handles data splitting into training, validation, and test sets. 
It also defines the data input and output dimensions, which can be used to modify the \texttt{BaseModel} by the \texttt{BaseMethod}.

\subsection{\texttt{BaseModel}}\label{sec:core-components-modules:model}

The \texttt{BaseModel}, defined in \path{yamle/models/model.py}, is responsible for determining the architecture of the model and its forward pass. 
It defines several components, such as the \texttt{\_input} and \texttt{\_output} layers, which can be modified by the \texttt{BaseMethod}. 
The goal is to write general and configurable implementations of a model that can be used across different datasets and tasks.
For example, if defining a multi-layer perceptron, the \texttt{BaseModel} should be configurable to different widths, depths, and activation functions.

\subsection{\texttt{BaseMethod}}\label{sec:core-components-modules:method}

The \texttt{BaseMethod}, defined in \path{yamle/methods/method.py}, defines the interface that can optionally change the model and specifies the training, validation, and test steps by reusing \texttt{PyTorch Lightning}'s functionality. 
For instance, it can be used to implement a new training algorithm by overloading the \texttt{\_training\_step(**kwargs)}, \texttt{\_validation\_step(**kwargs)}, and \texttt{\_test\_step(**kwargs)} methods. 
Depending on the provided \texttt{BaseDataModule,} it automatically decides which are relevant algorithmic metrics to log and automatically logs them using callbacks provided by \texttt{PyTorch Lightning} at the end of each epoch.
The validation metrics are automatically passed to \texttt{syne-tune} for hyperparameter optimisation if it is desired.
The \texttt{BaseMethod} also considers the loss function, optimiser, and regularisation during training and can incorporate \texttt{Pruning} and \texttt{Quantisation} during evaluation. 

All the components—\texttt{BaseDataModule}, \texttt{BaseModel}, and \texttt{BaseMethod}—enable customisation by defining their arguments that can be triggered via \texttt{argparse}. 
These components are orchestrated by the \texttt{BaseTrainer/BaseTester}, responsible for querying the \texttt{BaseDataModule}, \texttt{BaseModel}, and \texttt{BaseMethod}, and executing training and evaluation loops through the step methods, as well as running on a specific device platform. 
These classes are connected and facilitate end-to-end experiments from data preprocessing to model training and evaluation. 
It only requires subclassing the appropriate classes, registering them in the framework for selection via \texttt{argparse}, and executing training or evaluation using the methods defined in \path{yamle/cli}.

\section{Use Cases and Applications}\label{sec:use-cases-applications}

\texttt{YAMLE} is designed to serve as the template for the main project itself rather than being used as an add-on to an existing project. 
The goal is to grow organically into a shared ecosystem where users can quickly build on and compare existing implementations without implementing the boilerplate code.
This can be accomplished through the following workflow:

\begin{enumerate}[leftmargin=*, topsep=0cm, itemsep=-0.1cm]
    \item The user clones the \texttt{YAMLE} repository and installs the project \texttt{pip install -e .}
    \item The user experiments with the new method or model by subclassing the \texttt{BaseMethod} or \texttt{BaseModel} on the chosen \texttt{BaseDataModule} or any other customisable component.
    \item Once the user is satisfied with their addition, e.g. they publish a paper or the feature is well received by the community, they add it to the repository by creating a pull request.
    \item During the pull request review, the feature will be added and categorised as a staple or an experimental feature.
    After new additions, a  new release of \texttt{YAMLE} will be distributed. 
\end{enumerate}

\noindent
\texttt{YAMLE} facilitates three primary use cases at the moment:

\begin{itemize}[leftmargin=*, topsep=0cm, itemsep=-0.1cm]
    \item \textbf{Training}: Users can initiate training using the command \texttt{python yamle/cli/train.py}, for example, with the following parameters:

    \texttt{python3 yamle/cli/train.py --method base --trainer\_devices "[0]" --datamodule mnist --datamodule\_batch\_size 256 --method\_optimizer adam --method\_learning\_rate 3e-4 --regularizer l2 --method\_regularizer\_weight 1e-5 --loss crossentropy  --save\_path ./experiments --trainer\_epochs 3 --model fc --model\_hidden\_dim 32 --model\_depth 3 --datamodule\_validation\_portion 0.1 --save\_path ./experiments}
    
    \item \textbf{Testing}: Users can perform testing by running \texttt{python yamle/cli/test.py}, for instance, with the following command:

    \item \texttt{python3 yamle/cli/test.py --method base --trainer\_devices "[0]" --datamodule mnist --datamodule\_batch\_size 256 --save\_path ./experiments --model fc --model\_hidden\_dim 32 --model\_depth 3 --datamodule\_validation\_portion 0.1 --load\_path ./experiments/<FOLDER>}

    \item \textbf{Hyperparameter Optimisation}: Users can optimise hyperparameters using the command \texttt{python yamle/cli/tune.py}, as shown in the following example:
    
    \texttt{python3 yamle/cli/tune.py --config\_file <FILE.py> --optimizer "Grid Search" --save\_path ./experiments/hpo/ --max\_wallclock\_time 420 --optimization\_metric "validation\_nll"}

\end{itemize}

Users can easily invoke training, testing, and hyperparameter optimisation for their models or methods, covering the entire pipeline from data preprocessing to model training and evaluation.

\section{Conclusion}\label{sec:conclusion}

\texttt{YAMLE} aims to be a one-stop-shop for ML experiments, enabling rapid prototyping and experimentation with ML models and methods, which can be easily extended and customised by modifying the core components and then used across different models, methods, datasets, and tasks.

\section*{Acknowledgements}

Martin Ferianc was sponsored through a scholarship from the Institute of Communications and Connected Systems at UCL. 
We thank Jaromir Latal, Kristina Ulicna, Ondrej Bohdal, and Afroditi Papadaki for their feedback on the draft. 
Martin would like to specifically thank Martin Wistuba, Giovanni Zappella, Lukas Balles, Gianluca Detommasso and the \texttt{syne-tune} team for providing feedback and suggestions on his coding style and allowing him to learn how to design and implement a complex ML project while at Amazon.

\bibliography{sample}

\appendix

\section{Related Work}\label{sec:related-work}

The ML landscape is rich with various libraries and frameworks, each serving specific roles in the research and development process. 
This Section examines several key projects in this space and highlights how \texttt{YAMLE}: Yet Another Machine Learning Environment sets itself apart.

\subsection{Core Machine Learning Libraries}\label{sec:related-work:core-ml-libraries}

\texttt{PyTorch}~\citep{paszke2017automatic}, \texttt{TensorFlow}~\citep{45381}, and \texttt{Jax}~\citep{frostig2018compiling} provide the fundamental building blocks for developing and training ML models. 
These core libraries offer atomic functionality for defining models, training algorithms, and evaluation metrics.
However, they do not provide a complete end-to-end solution for conducting experiments, requiring users to integrate the atomic operations into larger pipelines or higher-level frameworks.
At the same time, they are not designed as the design space for a project but as the main building blocks.

\subsection{Higher-Level Frameworks}\label{sec:related-work:higher-level-frameworks}

Higher-level frameworks like \texttt{PyTorch Lightning}~\citep{falcon2019pytorch} and \texttt{Keras}~\citep{chollet2015keras} aim to simplify the training and evaluation of ML models. 
While both offer a standardised interface for training, optimisation, and evaluation, users must still orchestrate the connections between data, models, and methods. It also lacks a comprehensive collection of pre-implemented models, techniques, and datasets for quick and direct model comparisons.
Therefore, researchers must reimplement the same boilerplate code, which can lead to disparate implementations and a lack of standardisation, hindering reproducibility and comparison of results~\citep {semmelrock2023reproducibility}.

\subsection{Domain-Specific Libraries}\label{sec:related-work:domain-specific-libraries}

Several domain-specific libraries like \texttt{Edward2}~\citep{tran2018simple}, \texttt{Fortuna}~\citep{detommaso2023fortuna}, \texttt{Pyro}~\citep{bingham2018pyro}, \texttt{Renate}~\citep{wistuba2023renate}, \texttt{Albumentations}~\citep{buslaev2020albumentations}, and \texttt{AllenNLP}~\citep{gardner2018allennlp} offer unified interfaces for specific ML areas. 
These libraries excel in their respective domains but are typically used as additional components within a project rather than serving as their primary framework.
\texttt{Ludwig}~\citep{molino2019ludwig} provides a high-level interface for training and evaluating deep learning models, making it easier for users to specify data and model architectures using declarative language. 
It automates tasks such as preprocessing, feature extraction, and hyperparameter optimisation.
However, expanding its capabilities can make it more complex and less flexible than \texttt{YAMLE}, which focuses on rapid prototyping and experimentation.
\texttt{MLflow}~\citep{chen2020developments}, on the other hand, offers tools for managing the entire ML lifecycle, with a focus on deployment and production readiness. 
In contrast, \texttt{YAMLE} is designed to facilitate rapid prototyping and experimentation, with a unique emphasis on simplifying the research and development phase.

\subsection{Comparison with \texttt{YAMLE}}\label{sec:related-work:comparison-with-yamle}

\texttt{YAMLE} stands out as an environment designed to provide a comprehensive end-to-end solution for ML experimentation.
Unlike many existing libraries, it offers a modular architecture encompassing data, models, and methods, allowing users to customise these components to meet their needs. 
Notably, \texttt{YAMLE} is engineered to be the central framework for a project rather than just an add-on, providing a unified environment for conducting experiments.

One of the key differentiators is \texttt{YAMLE}'s ambition to create a shared ecosystem where researchers and practitioners can efficiently build on and compare existing implementations. 
This is achieved by offering a growing collection of models, methods, and datasets for direct model-to-model and method-to-method comparisons. 
In this way, \texttt{YAMLE} aims to be the cornerstone of a broader community where researchers use the environment to assess their methods and models on the same datasets and tasks. 
When they publish their work, they can seamlessly add their methods or models to the \texttt{YAMLE} repository for others to use and compare.

In summary, while existing libraries and frameworks cover various aspects of the ML pipeline, \texttt{YAMLE} distinguishes itself by providing an integrated environment for rapid experimentation, minimising the barriers to reuse and extension. 
By offering modularity, a comprehensive ecosystem, and the main project frameworks, \texttt{YAMLE} opens the door to more accessible and efficient ML research.

\section{Future Development}\label{sec:future-development-roadmap}

The currently implemented methods and tasks focus mainly on supervised regression and classification tasks.
For the author's interests, there is also a focus on uncertainty quantification methods paired with out-of-distribution detection methods.
The project is still in its infancy, with many areas for improvement and extension.

\begin{itemize}[leftmargin=*, topsep=0cm, itemsep=-0.1cm]
    \item \textbf{Documentation}: The environment currently lacks extensive documentation, a priority for future development.
    \item \textbf{Additional Tasks}: The environment would greatly benefit from other problems common in ML, such as unsupervised or self-supervised learning or reinforcement learning.
    \item \textbf{Expanding the BaseModel Zoo}: The environment would greatly benefit from a larger collection of models and methods to compare the new model or method to the existing ones.
    \item \textbf{Testing}: The environment is currently lacking unit tests, and it relies on the correctness of the \texttt{PyTorch Lightning}, \texttt{torchmetrics}, and \texttt{syne-tune} libraries for the pipeline flow, metrics, and hyperparameter optimisation respectively.
    \item \textbf{Multi-device Runs}: The environment currently supports training/testing on a single device, but it would be beneficial to support multi-device usage.
    \item \textbf{Other Hyperparameter Optimisation Methods}: The environment currently supports \texttt{syne-tune} for hyperparameter optimisation, but it would be beneficial to support other methods such as \texttt{Optuna}~\citep{akiba2019optuna} or \texttt{Ray Tune}~\citep{liaw2018tune}.
\end{itemize}

\end{document}